\documentclass{article}
\usepackage{microtype}
\usepackage{graphicx}
\usepackage{booktabs}
\usepackage{multirow}
\usepackage{xcolor}
\usepackage{colortbl}
\usepackage{hyperref}

\usepackage[accepted]{icml2025}
\makeatletter\renewcommand{\ICML@appearing}{Preprint.}\renewcommand{\Notice@String}{Preprint.}\makeatother
\usepackage{amsmath}
\usepackage{amssymb}
\usepackage{amsthm}
\usepackage{bm}
\usepackage{pifont}
\usepackage{enumitem}
\usepackage{xspace}
\usepackage{inconsolata}   %
\usepackage[capitalize,noabbrev]{cleveref}

\definecolor{readc}{HTML}{9A6B0E}
\definecolor{saidc}{HTML}{2E4A72}
\definecolor{oursbg}{HTML}{FBF4E2}
\definecolor{refc}{gray}{0.45}
\definecolor{headbg}{HTML}{F1EFEA}
\definecolor{gainc}{HTML}{D9481C}
\newcommand{\oursrow}{\rowcolor{oursbg}}
\newcommand{\headrow}{\rowcolor{headbg}}
\newcommand{\gain}[1]{\textsubscript{\textcolor{gainc}{$\uparrow$#1}}}
\newlength{\gainw}

\newcommand{\ours}{\mbox{\textsc{Legible}}\xspace}
\newcommand{\obs}[1]{\textbf{Obs.~\ding{\numexpr191+#1\relax}}}
\newcommand{\bacc}{\mbox{B-Acc}\xspace}
\makeatletter\newcommand{\figcaption}[1]{\def\@captype{figure}\caption{#1}}\makeatother

\newcommand{\teasercaption}{\textbf{A fine-tuned fact can be said but not used.} (1) Fine-tuned on a new fact, the model says it, yet answers every two-hop question about Lena Hart for her old country. (2) The second hop reads its bridge in the middle layers (light gold), still as the old country; fine-tuning writes near the output (blue). (3) Existing fixes write a word to be said (blue); \ours writes the entity as the model reads it (dark gold). (4) Learned facts then compose more often than the model's own knowledge (Qwen2.5-7B; \S\ref{sec:main}). Illustrative example.}
\icmltitlerunning{Write Facts the Way Models Read Them}

\begin{document}

\twocolumn[
\icmltitle{Write Facts the Way Models Read Them: Lifting the Two-Hop Curse}

\begin{icmlauthorlist}
\icmlauthor{Zhiyun Shi}{ntu}
\end{icmlauthorlist}
\icmlaffiliation{ntu}{Nanyang Technological University, Singapore}
\icmlcorrespondingauthor{Zhiyun Shi}{ZHIYUN004@e.ntu.edu.sg}
\icmlkeywords{knowledge editing, two-hop curse, latent multi-hop reasoning, interpretability}
\vskip 0.15in
\begin{minipage}{\textwidth}
\centering
{\includegraphics[width=\textwidth]{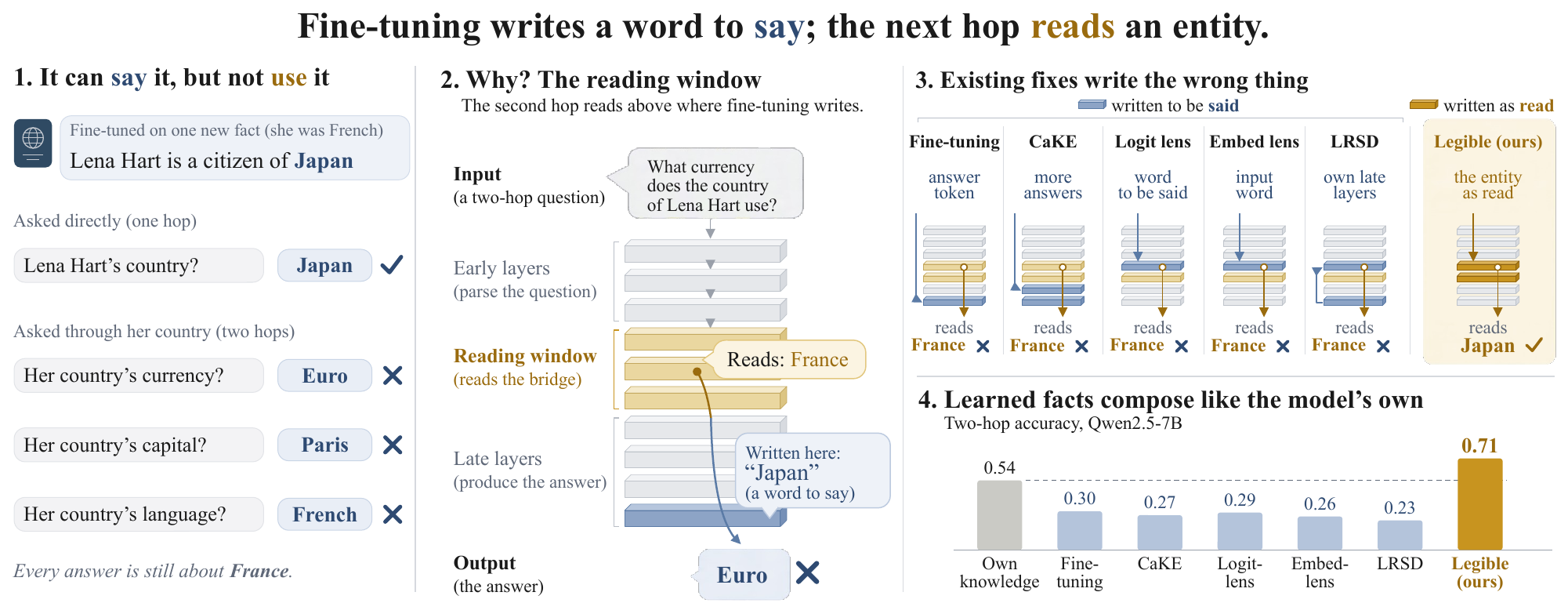}}
\setlength{\abovecaptionskip}{0pt}\vspace{-12pt}
\figcaption{\teasercaption}
\label{fig:teaser}
\end{minipage}
\vskip 0.2in
]
\printAffiliationsAndNotice{}

\begin{abstract}
A language model that learns a new fact by fine-tuning can repeat it but cannot reason with it: taught that a person is a citizen of a new country, it still cannot say, without chain-of-thought, who the head of state of that person's country is.
This \textbf{two-hop curse} is widely treated as a reasoning deficit, and the strongest knowledge editors answer it with synthesized reasoning data.
\textbf{We show that the reasoning is intact and the writing is wrong.}
Fine-tuning writes a new fact as an \emph{answer to be said}, whereas the second hop reads the bridge (``that person's country'') as an \emph{entity}, from the middle layers on: patching the model's own representation of the entity into the bridge at those layers largely restores the second hop, yet after fine-tuning the bridge there is no closer to the new entity than before the edit.
We therefore propose \textbf{read-equivalent \mbox{writing}} (\ours): one loss term makes the bridge phrase, at the layer where the second hop finishes reading it, match the model's own reading of the new entity, and keeps the same relation's other subjects unchanged.
Phrases and targets come from the edited model itself, and no multi-hop training data is used.
Across Qwen2.5, \mbox{Llama-3} and Qwen3, \ours
\textbf{(I)} composes learned facts \textbf{more often than the model composes its own knowledge}, on MQuAKE's two-hop-curse questions; %
\textbf{(II)} reaches \textbf{1.8--2.6$\boldsymbol{\times}$} the bridge accuracy of CaKE, whose reasoning data composes no better than plain fine-tuning, and has the best average over composition, accuracy and locality against four families of writers; %
and \textbf{(III)} lifts the curse on RippleEdits, on the benchmark of Balesni et al., who named the curse, and at 14B.
\end{abstract}
\section{Introduction}\label{sec:intro}

In Plato's \emph{Phaedrus}, Socrates complains that written words seem to think, yet ask them anything and they only repeat the same thing~\citep{plato_phaedrus}.
Facts that a language model learns by fine-tuning behave the same way.
Taught that a person is now a citizen of Japan, the model answers ``Which country is she a citizen of?'' correctly, but asked ``Who is the head of state of the country she is a citizen of?'' it names the head of state of her old country, or of no country at all.
The community has several names for this failure: the \textbf{two-hop curse}~\citep{balesni}, ripple effects~\citep{rippleedits}, multi-hop editing failures~\citep{mquake}, and the knowing--using gap~\citep{lrsd}.
They describe one phenomenon: a learned fact can be stated but not used.

The prevailing reading is that the model has not learned to \emph{reason} with the new fact, and the remedy is more reasoning-shaped supervision.
CaKE~\citep{cake}, the strongest parametric editor, asks an external LLM to write circuit-aware practice questions for every edit; \citet{kureason} state that knowledge updating is a reasoning problem and distil the model on self-generated multi-hop questions; \citet{balesni} add auxiliary losses at the bridge, and LRSD~\citep{lrsd} pulls the subject's middle layers towards its own late layers.
All of these assume the missing piece is reasoning.

\textbf{The reasoning is intact; the writing is wrong.}
Each hop alone works: after editing, the new fact is recalled 97--100\% of the time and the base model knows 86--91\% of the second hops. %
On the same chains built from its own knowledge, the base model composes the two hops without chain-of-thought about half of the time (51--55\% on the three models). %
What breaks is how the new fact is written, in knowledge editing, where the bridge is an entity the model already knows.
The second hop reads its bridge entity in the middle layers: patching the model's own representation of the entity into the bridge position at those layers largely restores the second hop, while patching at late layers does not.
After fine-tuning, the bridge at those layers is no closer to the new entity than before the edit, with plain fine-tuning and CaKE alike (\S\ref{sec:diagnosis}). %
Fine-tuning writes the new fact as an answer to be said; the second hop needs it as an entity to be read, and where it reads, the fine-tuned bridge never becomes it.

This points to a principle rather than a module.
A write must produce the representation that downstream computation reads; we call this the \textbf{read--write contract}.
\ours writes each fact the way the model reads it: at the layer where the second hop finishes reading it, the bridge phrase (``the country of citizenship of $s$'') is pulled to the model's own state when it reads ``Japan'' in a neutral sentence.
A second term keeps the relation's other subjects where they were, so that an edit changes one value of one function.
Phrases and targets come from the edited model itself; no multi-hop question, external LLM or template is used.

\textbf{Every existing remedy is a choice of what to write the bridge as.}
Reasoning data still writes with cross-entropy, so it writes more answers: on the 213 two-hop-curse questions CaKE is indistinguishable from plain paraphrase fine-tuning on all three models (0.27--0.34 vs.\ 0.30), and its gains lie entirely on questions whose answer was already written in its training text. %
The direct fixes write the wrong thing at the right place: Balesni's logit-lens loss writes the token to be said, the embed-lens loss the input word, and LRSD the model's own late layers.
None of the fine-tuning writers moves the bridge towards the new entity inside the reading window, and none composes better than plain fine-tuning. The one writer that does move it, AlphaEdit, never targets the bridge: it writes the fact where the model keeps its own facts, and composes as far as it moves the bridge (\S\ref{sec:analysis}).

The contract also predicts where writing as read helps and where it cannot.
\emph{Where it helps}: questions whose answer is never the object being written, which is the two-hop curse itself.
On the 213 such questions in MQuAKE, learned facts compose more often than the model's own knowledge, about twice as often as with CaKE, and the curse lifts on RippleEdits and on Balesni's semi-synthetic benchmark as well.
\emph{Where it cannot}: questions whose answer is the written word itself, where plain fine-tuning is already at 0.81--0.89. %
On the standard M-Acc, which mixes both kinds, our two variants score highest on all three models, \ours has the best average over accuracy, composition and locality, and the gain holds at 14B.

Our contributions are as follows:
\begin{itemize}[leftmargin=*,itemsep=2pt,topsep=2pt]
\item \textbf{Paradigm Reformulation.} We recast the two-hop curse as a write-time failure, state the read--write contract, and show that reasoning data, lens-based losses and self-distillation are special cases that write the bridge as something other than what the model reads.
\item \textbf{Practical Solution.} \ours, a single loss that writes a fact the way the edited model reads it, with phrases and targets produced by the model itself, no multi-hop training data, and the best locality among fine-tuning writers.
\item \textbf{Experimental Evaluation.} On MQuAKE, RippleEdits and Balesni's semi-synthetic benchmark with three LLMs, learned facts compose more often than the model's own knowledge on MQuAKE (1.8--2.6$\times$ CaKE on bridge accuracy), and at 14B; across four families of writers, composition follows what the bridge reads as.
\end{itemize}

\begin{figure*}[t]
\centering
\includegraphics[width=\textwidth]{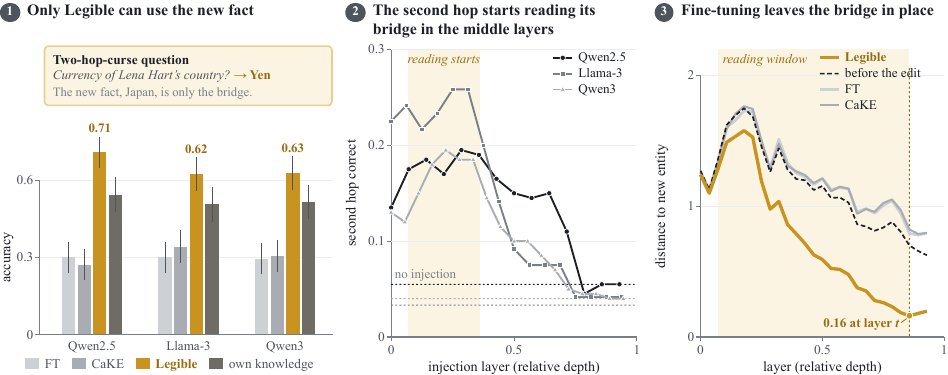}
\caption{\textbf{Fine-tuning writes the bridge as an answer, not as an entity the second hop can read.} (1) On the 213 two-hop-curse questions per model, fine-tuning and CaKE answer about 0.3, while \ours exceeds the base model's accuracy with its own knowledge on the same chains (bars: 95\% intervals). (2) Injecting the base model's reading of the true bridge entity at one layer restores the second hop only by the middle layers, where reading starts (held-out chains, base model; dotted: no injection). (3) At the bridge of the test question (Qwen2.5, medians over the 154 two-hop-curse questions whose hidden states were recorded), fine-tuning and CaKE leave the bridge as far from the new entity as before the edit; \ours moves it from layer 8 on and brings it there by layer $t$, where reading ends. The example question is illustrative.}
\label{fig:diag}
\end{figure*}
\section{The Two-Hop Curse Is a Write-Time Failure}\label{sec:diagnosis}

\subsection{Two kinds of questions}\label{sec:kinds}

A multi-hop question \mbox{$s \xrightarrow{r_1} o_1 \xrightarrow{r_2} o_2$} chains single-hop facts.
After a fact is edited, two kinds of question arise.
If the edited hop is the last one, the bridge is an entity the model already knew, and the answer is the newly written object itself, the very text fine-tuning trains the model to say.
If an edited hop is followed by an unedited one, the new object $o_1^\ast$ is only a \emph{bridge}: the answer is a fact the model already knew about $o_1^\ast$ and is never the object being written.
We call the second kind \textbf{two-hop-curse questions}.
This is the setting of knowledge editing, and of the semi-synthetic benchmark of \citet{balesni}, who named the curse.

We report three accuracies.
\textbf{M-Acc} is the standard multi-hop accuracy of CaKE's protocol~\citep{cake}.
\textbf{\bacc} (bridge accuracy) is the accuracy on two-hop-curse questions.
\textbf{CR} (composition retention) is \bacc restricted to chains that the base model composes correctly with its original facts: of the compositions the model can do with its own knowledge, how many survive when one hop is written by fine-tuning.
All three are read against the base model's accuracy on the original chains: its own knowledge.

In MQuAKE-CF-3k-v2, two-hop questions that end in the written object dominate: 241 of the 300 Qwen test cases, where plain fine-tuning is already correct \mbox{0.81--0.89} of the time. %
On two-hop-curse questions, fine-tuning and CaKE answer about 0.3 on all three models (Figure~\ref{fig:diag}, panel~1).

\subsection{Where the second hop reads its bridge}\label{sec:window}

To locate where the second hop reads its bridge, we take a two-hop question, write the first-hop description (``the country of citizenship of $s$'') into the second-hop prompt, and at its last token replace the hidden state at layer $t$ with the base model's own representation of the true bridge entity $o$, read from the neutral sentence ``This is about $o$''.
No training is involved.
On held-out MQuAKE chains, injection in the early-to-middle layers (layers 2--10 for Qwen2.5-7B, 6--12 for Qwen3-8B) lifts the second hop from 0.03--0.06 to 0.20--0.26, 60--90\% of what naming the entity in the question gives (0.22--0.34), and fails at late layers. %
There is a \textbf{reading window}: the second hop starts reading its bridge in the middle layers, so an entity injected later arrives too late (Figure~\ref{fig:diag}, panel~2), and it keeps reading until late, layer 24 of 28 on Qwen2.5 (cutting attention to the bridge only from there on leaves the answers intact; Appendix~\ref{app:inject}). %

\subsection{Fine-tuning writes the bridge as an answer}\label{sec:answer}

We now look at the bridge position of the test question itself, after a fact has been written by fine-tuning.
In \mbox{25--56\%} of their errors on two-hop-curse questions, fine-tuning and CaKE answer with the written bridge itself (``Japan'' for a head of state).
On MQuAKE, we measure at every layer the relative squared distance between the bridge's hidden state in the test question and the base model's read representation of the new entity.
After plain fine-tuning or CaKE, the bridge is about as far from the new entity as it was before the edit at every layer of the window, and slightly farther where reading ends (layer 24: 0.79 and 0.82, against 0.70 before the edit), whereas \ours brings it to 0.16 (Figure~\ref{fig:diag}, panel~3). %
Contrasting the two entities shows what the fine-tuned bridge reads as: at layer 24 the bridge is closer to the new entity than to the old one in 23\% of the questions before the edit, and still in only 38\% after fine-tuning and 37\% after CaKE; after \ours it is closer to the new entity in every question. %
The fine-tuned model states the new fact, yet the second hop reads the old entity.

Repairing the bridge alone repairs the composition.
On the cloze form of the 213 two-hop-curse questions, replacing the bridge's hidden state at a window layer with the base model's read representation of the new entity, changing nothing else, lifts every baseline writer from \mbox{0.21--0.28} to \mbox{0.54--0.55} on Qwen2.5, 0.35--0.45 on Llama-3 and \mbox{0.45--0.51} on Qwen3; the old entity or an unrelated one in the same place gives at most 0.20.
Conversely, in models written as read (aligned at that window layer), putting the old entity back into the bridge takes composition from 0.51--0.53 to \mbox{0.12--0.31}. %
In the unedited model the same replacement gives 0.68, 0.57 and 0.58, more than the model composes with its own facts on this form (0.51, 0.43, 0.45): a bridge read as the entity is the ceiling of composition, and \ours reaches it (0.63, 0.56, 0.58; \S\ref{sec:main}). %
The second hop of the fine-tuned model is intact; only its bridge is wrong.

These observations fix what a write must do.
\begin{quote}
\textbf{Read--write contract.} A write must produce the representation that downstream computation reads: the new fact, at the bridge, must look the way the model looks when it reads the new entity, until the second hop has finished reading it.
\end{quote}
\section{Writing Facts the Way the Model Reads Them}\label{sec:method}

\begin{figure*}[t]
\centering
\includegraphics[width=\textwidth]{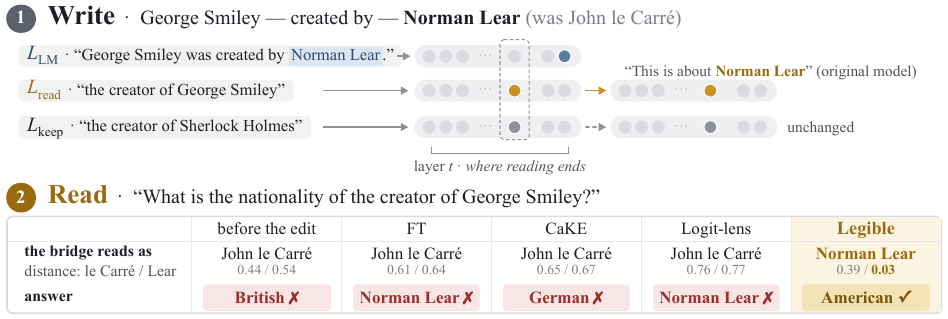}
\caption{\textbf{\ours on one test question} (MQuAKE, Qwen2.5-7B). (1)~\emph{Write.} $\mathcal{L}_{\text{LM}}$ trains the edit sentence, a word to say at the last layer; $\mathcal{L}_{\text{read}}$ pulls the bridge of a description written by the model, at layer $t=24$, where the second hop finishes reading the bridge, to how the original model reads ``Norman Lear''; $\mathcal{L}_{\text{keep}}$ holds other subjects of the relation where they were. (2)~\emph{Read.} At the bridge of the two-hop question, layer 24, every other writer leaves the bridge closer to the old entity (distances to John le Carr\'e / Norman Lear), including the two that answer ``Norman Lear''; only \ours makes it the new entity, and only \ours answers correctly.}
\label{fig:method}
\end{figure*}

\subsection{Read-equivalent writing}\label{sec:readeq}

Let a fact $(s, r, o)$ be written into a model $f_\theta$ with parameters $\theta$, starting from the base model $f_{\theta_0}$.
The write is trained on the usual language-modelling loss $\mathcal{L}_{\text{LM}}$ over the edit sentence and its paraphrases.
For the bridge, we take a few descriptions $d(s)$ of the edited relation that end at the bridge, such as ``\emph{the country of citizenship of $s$}'' and ``\emph{$s$'s country of citizenship}'', and denote by $h_t(d(s))$ the hidden state of the edited model at their last token and layer $t$.
The target is how the base model reads the new entity:
\begin{equation}
e_t(o) = h^{\theta_0}_t\big(\text{``This is about } o\text{''}\big)\big|_{\text{last token of } o}.
\end{equation}
Read-equivalent writing adds one term,
\begin{equation}
\mathcal{L}_{\text{read}} = \frac{\lVert h_t(d(s)) - e_t(o)\rVert^2}{\lVert e_t(o)\rVert^2},
\end{equation}
averaged over the descriptions.
The target is neither the answer token nor the input embedding of $o$: it is the state the model is in when it has \emph{read} $o$, at the same layer (Figure~\ref{fig:method}).

\subsection{One edit, one function value}\label{sec:column}

An edit changes the value of the function $r$ at the point $s$.
Read-equivalent writing fixes that point; the rest of the function should not move.
We therefore keep, for a few other subjects $s'$ of the same relation, the bridge exactly where the base model had it:
\begin{equation}
\mathcal{L}_{\text{keep}} = \frac{1}{|S'|}\sum_{s'\in S'} \frac{\lVert h_t(d(s')) - h^{\theta_0}_t(d(s'))\rVert^2}{\lVert h^{\theta_0}_t(d(s'))\rVert^2}.
\end{equation}
The full objective is
\begin{equation}
\mathcal{L} = \mathcal{L}_{\text{LM}} + w\,(\mathcal{L}_{\text{read}} + \mathcal{L}_{\text{keep}}).
\end{equation}
Together the two terms say what an edit is: the edited cell reads as $o$, and the rest of its column reads as before.
We call the result \ours: a fact is legible when the model can read it, not only say it.

\subsection{Everything comes from the model}\label{sec:self}

\ours needs no external resource.
The \emph{descriptions} are produced by the edited model: given only the single-hop template of a relation (``The capital of \{\} is''), it names the role of the answer (``capital''), from which we form ``the capital of $s$'' and ``$s$'s capital''; the few-shot examples use relations outside the benchmarks. %
The \emph{target} is the model's own reading of $o$.
The \emph{layer} $t$ is where the second hop finishes reading its bridge, chosen on chains outside the test set by one rule for all models: the shallowest layer from which cutting all attention to the bridge still leaves at least 90\% of the base model's correct answers ($t=24$ for Qwen2.5-7B, $18$ for Llama-3-8B, $26$ for Qwen3-8B; Appendix~\ref{app:inject}). %
The injection plateau of \S\ref{sec:window} marks where reading starts; aligning where it ends keeps the bridge the new entity for as long as it is read.
The \emph{neighbours} $s'$ are three subjects of the same relation that occur nowhere in the evaluation data, its locality probes included.
No multi-hop question, external LLM or relation-specific template enters training.

\subsection{Existing writers as special cases}\label{sec:special}

Every method we compare to is the same objective with a different answer to one question: what should the bridge become?
\begin{equation}
\mathcal{L} = \mathcal{L}_{\text{LM}} + w\cdot D\big(h(\text{bridge}),\ \text{target}\big).
\end{equation}
Table~\ref{tab:special} lists the choices.
Plain fine-tuning and CaKE set no target at the bridge, so what is written there is shaped by the answer; CaKE's reasoning text is written with cross-entropy too, as more answers.
Balesni's lens losses target the token to be \emph{said} or the \emph{input} word, LRSD the subject's own late layers, and self-distillation~\citep{kureason} a teacher that has the new fact in context.
Only \ours targets what the second hop reads.

\begin{table}[t]
\centering
\footnotesize
\setlength{\aboverulesep}{0pt}\setlength{\belowrulesep}{0pt}\renewcommand{\arraystretch}{1.1}
\setlength{\tabcolsep}{3pt}
\begin{tabular}{lll}
\toprule
\headrow \textbf{Writer} & \textbf{Target at the bridge} & \textbf{Multi-hop data} \\
\midrule
Fine-tuning & none (answer) & no \\
CaKE & none (more answers) & LLM-written \\
Embed-lens & input embedding of $o$ & no \\
Logit-lens & token $o$ to be said & no \\
LRSD & own late layer, at $s$ & no \\
Self-distillation & teacher with context & self-generated \\
\oursrow \textbf{\ours} & \textcolor{readc}{\textbf{how the model reads $o$}} & no \\
\bottomrule
\end{tabular}
\caption{Every writer is the same objective with a different target at the bridge: fine-tuning, CaKE~\citep{cake}, the two auxiliary losses of \citet{balesni}, LRSD~\citep{lrsd}, self-distillation~\citep{kureason}, and \ours.}
\label{tab:special}
\end{table}
\section{Experiments}\label{sec:experiments}

\begin{table*}[t]
\centering
\footnotesize
\setlength{\aboverulesep}{0pt}\setlength{\belowrulesep}{0pt}\renewcommand{\arraystretch}{1.1}
\setlength{\tabcolsep}{1.3pt}
\begin{tabular}{l r r r >{\hspace{3pt}}r r >{\hspace{3pt}}r r r r >{\hspace{3pt}}r r >{\hspace{3pt}}r r r r >{\hspace{3pt}}r r >{\hspace{3pt}}r}
\toprule
\headrow & \multicolumn{6}{c}{\textbf{Qwen2.5-7B}} & \multicolumn{6}{c}{\textbf{Llama-3-8B}} & \multicolumn{6}{c}{\textbf{Qwen3-8B}} \\
\cmidrule(lr){2-7}\cmidrule(lr){8-13}\cmidrule(lr){14-19}
\headrow \textbf{Method} & \multicolumn{1}{c}{M-Acc} & \multicolumn{1}{c}{\bacc} & \multicolumn{1}{c}{CR} & \multicolumn{1}{c}{Nbr} & \multicolumn{1}{c}{Unrel} & \multicolumn{1}{c}{Avg} & \multicolumn{1}{c}{M-Acc} & \multicolumn{1}{c}{\bacc} & \multicolumn{1}{c}{CR} & \multicolumn{1}{c}{Nbr} & \multicolumn{1}{c}{Unrel} & \multicolumn{1}{c}{Avg} & \multicolumn{1}{c}{M-Acc} & \multicolumn{1}{c}{\bacc} & \multicolumn{1}{c}{CR} & \multicolumn{1}{c}{Nbr} & \multicolumn{1}{c}{Unrel} & \multicolumn{1}{c}{Avg} \\
\midrule
\textcolor{refc}{\textit{Own knowledge}} & -- & \textcolor{refc}{0.54} & \textcolor{refc}{1.00} & -- & -- & -- & -- & \textcolor{refc}{0.51} & \textcolor{refc}{1.00} & -- & -- & -- & -- & \textcolor{refc}{0.52} & \textcolor{refc}{1.00} & -- & -- & -- \\  %
\midrule
FT & 0.71 & 0.30 & 0.37 & 0.42 & 0.83 & 0.56 & 0.68 & 0.30 & 0.35 & 0.26 & 0.65 & 0.47 & 0.67 & 0.30 & 0.37 & 0.33 & 0.75 & 0.51 \\  %
CaKE & 0.75 & 0.27 & 0.32 & 0.39 & 0.82 & 0.56 & 0.75 & 0.34 & 0.38 & 0.25 & 0.71 & 0.51 & 0.73 & 0.31 & 0.37 & 0.31 & 0.75 & 0.52 \\  %
\oursrow \quad + \ours & \textbf{0.87} & \underline{0.64} & \underline{0.75} & \underline{0.63} & \underline{0.90} & \underline{0.76} & \underline{0.79} & \textbf{0.63} & \textbf{0.71} & \underline{0.45} & \textbf{0.81} & \underline{0.67} & \textbf{0.80} & \underline{0.57} & \underline{0.69} & \underline{0.56} & \underline{0.87} & \underline{0.70} \\  %
\midrule
Self-distillation & 0.70 & 0.34 & 0.42 & 0.43 & 0.79 & 0.57 & 0.73 & 0.38 & 0.44 & 0.33 & 0.73 & 0.54 & 0.75 & 0.38 & 0.45 & 0.37 & 0.73 & 0.56 \\  %
BRACE$^\ddagger$ & 0.59 & 0.36 & 0.47 & 0.48 & 0.81 & 0.56 & 0.70 & 0.37 & 0.43 & 0.37 & 0.70 & 0.54 & 0.67 & 0.36 & 0.46 & 0.39 & 0.79 & 0.55 \\  %
\midrule
Embed-lens & 0.64 & 0.26 & 0.34 & 0.31 & 0.75 & 0.49 & 0.60 & 0.15 & 0.17 & 0.09 & 0.54 & 0.35 & 0.59 & 0.20 & 0.28 & 0.22 & 0.64 & 0.41 \\  %
Logit-lens & 0.72 & 0.29 & 0.37 & 0.41 & 0.82 & 0.56 & 0.68 & 0.26 & 0.31 & 0.16 & 0.57 & 0.42 & 0.64 & 0.30 & 0.38 & 0.38 & 0.80 & 0.53 \\  %
LRSD & 0.74 & 0.23 & 0.27 & 0.16 & 0.55 & 0.42 & 0.67 & 0.11 & 0.12 & 0.03 & 0.07 & 0.22 & 0.66 & 0.25 & 0.34 & 0.14 & 0.47 & 0.38 \\  %
\midrule
\oursrow \ours & \underline{0.83} & \textbf{0.71} & \textbf{0.81} & \textbf{0.66} & \textbf{0.92} & \textbf{0.78} & \textbf{0.83} & \underline{0.62} & \underline{0.68} & \textbf{0.46} & \underline{0.79} & \textbf{0.68} & \underline{0.79} & \textbf{0.63} & \textbf{0.78} & \textbf{0.62} & \textbf{0.92} & \textbf{0.74} \\  %
\bottomrule
\end{tabular}
\caption{Main results on MQuAKE. M-Acc: multi-hop accuracy under CaKE's protocol. \bacc: accuracy on the 213 two-hop-curse questions outside the development split. CR: \bacc on the chains the base model composes with its own knowledge. Nbr / Unrel: same-relation neighbours and unrelated facts the base model answered correctly. Avg: mean of M-Acc, \bacc, Nbr and Unrel. \emph{Own knowledge}: the base model on the original chains. Rows are grouped by family: fine-tuning, training data, bridge targets, ours; + \ours: CaKE trained with \ours added. AlphaEdit: Table~\ref{tab:ae}. $^\ddagger$BRACE generates 64 tokens as in its protocol and is scored on the first sentence; under the 10-token budget of all other rows its \bacc is 0.25 / 0.24 / 0.26, as it names a descriptor before each entity (Appendix~\ref{app:brace}). \textbf{Bold}: best; \underline{underline}: second best. On all three models \ours is above every baseline on \bacc and CR ($p<10^{-3}$) and on Nbr ($p<0.05$) by a paired sign test over cases; 95\% intervals in Table~\ref{tab:ci}.}
\label{tab:main}
\end{table*}

\subsection{Setup}\label{sec:setup}

\paragraph{Models and protocol.}
We use Qwen2.5-7B-Instruct, Llama-3-8B-Instruct and Qwen3-8B~\citep{qwen25,llama3,qwen3}.
All writers follow CaKE's official protocol~\citep{cake}: one LoRA per edit case (rank 8, $\alpha=32$, on all attention and MLP projections~\citep{lora}), 40 epochs, batch size 4, learning rate $10^{-4}$ with linear decay; evaluation is greedy decoding with alias substring matching, on both the cloze and the question form. %
\textbf{FT} trains on the edit sentence and six paraphrases; \textbf{CaKE} adds CaKE's released circuit-aware training data; \textbf{Self-distillation}~\citep{kureason} adds multi-hop questions the model writes about the new object and answers with the new fact in context (about five per edit). %
\textbf{BRACE}~\citep{brace} writes the bridge in every training sentence with a descriptor drawn from one shared pool (``the Japanese automotive manufacturer Toyota'') and adds known facts about the bridge decorated from the same pool; as in its protocol, the tail is never decorated, and descriptors or facts that contain the answer are removed (Appendix~\ref{app:brace}). %
\ours adds its loss to FT with $w=3$, four model-generated descriptions per relation and three neighbours per edit. %
\mbox{\textbf{CaKE + \ours}} adds it to CaKE.
\textbf{AlphaEdit}~\citep{alphaedit}, the locate-then-edit representative, writes the fact into the subject's early MLP layers under a null-space constraint (official implementation; Appendix~\ref{app:alphaedit}).
Design choices were fixed on a development split (the first 100 MQuAKE and 100 RippleEdits cases outside the test set).

\paragraph{Benchmarks.}
\textbf{MQuAKE-CF-3k-v2}~\citep{mquake}: two-hop cases (300 for Qwen2.5, 100 for Llama-3 and Qwen3), and, for \bacc and CR, the 213 two-hop-curse questions outside the development split. %
\textbf{RippleEdits}~\citep{rippleedits}: the 273 popular-split edits with compositional tests, filtered by the official condition queries; compositionality I consists entirely of two-hop-curse questions, while compositionality II asks for the written object itself, reached through another subject. %
\textbf{Balesni's semi-synthetic benchmark}~\citep{balesni}: 18 sets of 20 facts ``fictional person $\to$ real entity'', each set written into one LoRA, the second hop left to the model's own knowledge (\S\ref{sec:balesni}). %

\subsection{Main results}\label{sec:main}

\obs{1} \textbf{Learned facts compose more often than the model's own.}
On the 213 two-hop-curse questions, \ours exceeds the accuracy the base model has on the same chains with its original facts: 0.71 vs.\ 0.54 on Qwen2.5, 0.62 vs.\ 0.51 on Llama-3 and 0.63 vs.\ 0.52 on Qwen3, and on the cloze and the question form alike (Qwen2.5: 0.63 vs.\ 0.51 and 0.48 vs.\ 0.40). %
CR gives the paired version: of the chains the model composes with its own knowledge, \ours keeps 0.81 / 0.68 / 0.78, CaKE 0.32 / 0.38 / 0.37 (96--3, 70--10 and 72--3 against CaKE on \bacc). %
Under a strict score that also rejects any answer naming the bridge or the old answer, the gap widens (0.50 / 0.45 / 0.53 against at most 0.20 / 0.18 / 0.27), and \ours still exceeds the model's own knowledge scored the same way (0.35 / 0.30 / 0.37; Appendix~\ref{app:strict}). %

\obs{2} \textbf{Reasoning data does not repair composition.}
On two-hop-curse questions, FT and CaKE are indistinguishable on all three models (0.30 vs.\ 0.27, 0.30 vs.\ 0.34, 0.30~vs.~0.31). %
Self-distillation on the model's own multi-hop questions adds a few points (0.34, 0.38, 0.38) and stays below the model's own knowledge. %
Sharing bridge descriptions across hops (BRACE) lands in the same band, 0.36, 0.37 and 0.36 even on its own 64-token outputs, against 0.71, 0.62 and 0.63 for \ours (85--10, 64--10 and \mbox{66--8}); its bridge reads as the new entity in 42--48\% of the reading window, no more than after fine-tuning (46--51\%; \S\ref{sec:distance}). %
Writing more answers writes more answers.
Added to \ours, CaKE's data helps mainly the answer side: \mbox{CaKE + \ours} leads M-Acc on Qwen2.5 and Qwen3. %

\obs{3} \textbf{The gain appears where the theory says, on every benchmark.}
RippleEdits compositionality I is made entirely of two-hop-curse questions: \ours answers 0.63, 0.49 and 0.56 of them on the three models, against 0.31, 0.29 and 0.29 for CaKE (199--19, 152--25 and 176--20; Table~\ref{tab:ripple}). %

\obs{4} \textbf{Composition comes without the locality cost of fine-tuning.}
Among fine-tuning writers, \ours preserves the most same-relation neighbours and unrelated facts on all three models (Qwen2.5 0.66 / 0.92 vs.\ CaKE 0.39 / 0.82), and it has the best average over M-Acc, \bacc and both locality measures (0.78, 0.68, 0.74). %
AlphaEdit, whose edit is keyed to the named subject, preserves nearly all of them (0.93--0.99), and composes only as a first hop: when the subject of the edited fact is itself a latent bridge, it answers 0.02--0.12 of the questions, which caps its M-Acc at 0.10--0.19 and its \bacc at 0.46--0.53 (Table~\ref{tab:ae}). %

\begin{figure*}[t]
\centering
\includegraphics[width=\textwidth]{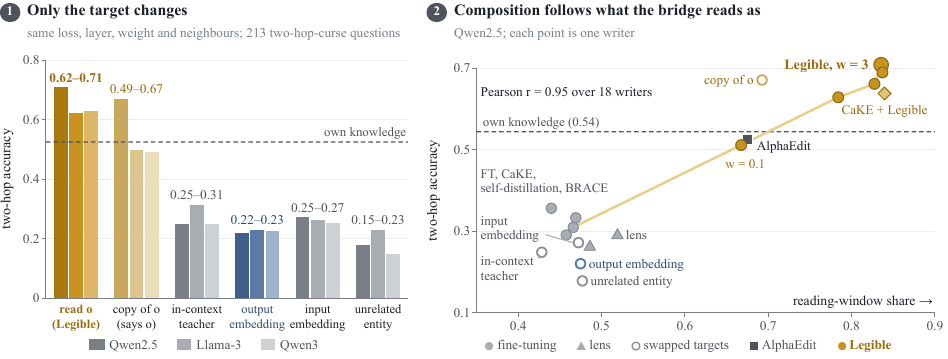}
\caption{\textbf{Composition follows what the bridge reads as, whichever writer.} (1) With the loss, layer, weight and neighbour term fixed, targets that read as $o$ compose: its reading (\ours) and a state that says $o$ by copying it; the model's state with the new fact in context, the output embedding of $o$, its input embedding and another entity's reading do not (dashed: the base model's own knowledge, 0.51--0.54). (2) For each writer on Qwen2.5, the share of the reading window (layers 2--24) in which the bridge of a test question is closer to the base model's reading of the new entity than to the old one, against accuracy on the 213 two-hop-curse questions; the gold line is \ours's weight sweep from plain fine-tuning ($w=0$) to $w=10$ ($w=3$ was fixed on the development split); open circles are the targets of (1).}
\label{fig:axis}
\end{figure*}

\subsection{Lifting the curse on its own benchmark}\label{sec:balesni}

\citet{balesni} named the two-hop curse on a benchmark built for it.
In its semi-synthetic version, fictional people are given a favourite or workplace real entity (``Which bridge does Zara Hughes-Marsjanska work at?''), and the two-hop question asks for a property of that entity the model already knows (``Consider the bridge where Zara Hughes-Marsjanska works. In which city is it located?''). %
The 18 sets give 1{,}380 two-hop questions; all are two-hop-curse questions, and all 20 facts of a set are written at once: knowledge injection, not single-fact editing.
We use the benchmark's own training conversations, prompts and scoring, and compare with fine-tuning and the authors' two auxiliary losses at the same position and layer.

On the curse's own benchmark, only writing as read lifts it (Table~\ref{tab:balesni}).
Plain fine-tuning reproduces the curse (0.17, as \citet{balesni} report), and the two auxiliary losses leave it in place (0.16--0.22), with their original coefficients or ours. %
\ours answers 0.49, against a ceiling of 0.67 when the real entity is named directly: it recovers about three quarters of the second hop the model can do, where fine-tuning recovers a quarter (460 wins vs.\ 29 losses), and it improves 17 of the 18 sets. %
On Qwen2.5: 0.11 for fine-tuning, 0.09--0.20 for the two auxiliary losses, and 0.41 for \ours against a ceiling of 0.66. %

\begin{table}[!h]
\centering
\footnotesize
\setlength{\aboverulesep}{0pt}\setlength{\belowrulesep}{0pt}\renewcommand{\arraystretch}{1.1}
\setlength{\tabcolsep}{5pt}
\begin{tabular}{l rr}
\toprule
\headrow \textbf{Writer} & \multicolumn{1}{c}{\textbf{Qwen2.5-7B}} & \multicolumn{1}{c}{\textbf{Llama-3-8B}} \\
\midrule
\textcolor{refc}{\textit{Ceiling (second hop)}} & \textcolor{refc}{0.66} & \textcolor{refc}{0.67} \\  %
\midrule
FT & 0.11 & 0.17 \\  %
Embed-lens & 0.11 & 0.17 \\  %
Logit-lens & \underline{0.20} & \underline{0.22} \\  %
\oursrow \ours & \textbf{0.41} & \textbf{0.49} \\  %
\bottomrule
\end{tabular}
\caption{Balesni's semi-synthetic benchmark: two-hop accuracy without chain-of-thought. Ceiling: the second hop asked with the real entity named. Lens losses: the better of their coefficient and ours (Table~\ref{tab:balfull}). \textbf{Bold}: best; \underline{underline}: second best.}
\label{tab:balesni}
\end{table}

\subsection{A larger model}\label{sec:scale}

\begin{table}[b]
\centering
\footnotesize
\setlength{\aboverulesep}{0pt}\setlength{\belowrulesep}{0pt}\renewcommand{\arraystretch}{1.1}
\setlength{\tabcolsep}{3.6pt}
\begin{tabular}{l rrrrr}
\toprule
\headrow \textbf{Qwen2.5-14B} & M-Acc & \bacc & CR & Nbr & Unrel \\
\midrule
\textcolor{refc}{\textit{Own knowledge}} & -- & \textcolor{refc}{0.61} & \textcolor{refc}{1.00} & -- & -- \\  %
\midrule
FT & 0.78 & 0.38 & 0.42 & \underline{0.36} & \underline{0.73} \\  %
CaKE & \underline{0.81} & \underline{0.42} & \underline{0.45} & 0.30 & 0.68 \\  %
\oursrow \ours & \textbf{0.87} & \textbf{0.78} & \textbf{0.79} & \textbf{0.56} & \textbf{0.78} \\  %
\bottomrule
\end{tabular}
\caption{Qwen2.5-14B, with the alignment layer chosen by the same rule ($t=36$ of 48) and no other change.}
\label{tab:q14}
\end{table}

At 14B, \ours is best on all five measures and again composes more often than the model's own knowledge: 0.78 against 0.61, CaKE 0.42 (83--5; Table~\ref{tab:q14}). %
Scale does not lift the curse for the baselines: from 7B to 14B, fine-tuning moves from 0.30 to 0.38 and CaKE from 0.27 to 0.42 on two-hop-curse questions, while \ours keeps 0.79 of the compositions the model can do with its own knowledge, against 0.42--0.45. %
\section{Why It Works: One Axis for Every Writer}\label{sec:analysis}

\subsection{Read versus said, with everything else fixed}\label{sec:ladder}

Table~\ref{tab:special} places all writers on one axis: what they write the bridge as.
We test the axis with everything but the target fixed: the same relative distance, layer $t$, weight $w=3$, descriptions and neighbour term.
The bridge is pulled towards the base model's reading of $o$ (\ours), towards the output embedding of $o$ rescaled to the same norm (the direction that makes the model say $o$), towards the input embedding of $o$, or towards the reading of an unrelated entity.
Only the reading of $o$ composes: 0.71, 0.62 and 0.63 on the three models, against 0.22--0.23 for the output embedding, 0.25--0.27 for the input embedding and 0.15--0.23 for another entity (Figure~\ref{fig:axis}, left). %
What matters is not whether the target says $o$ but whether it reads as $o$.
A state at layer $t$ that predicts $o$ by copying it (the last position of ``Name: $o$\texttt{\textbackslash n}Name: $o$\texttt{\textbackslash n}Name:'', where the model says $o$ 99--100\% of the time) carries the model's reading of $o$, and composes (0.67, 0.50, 0.49); the output embedding says $o$ without reading it, and composes below plain fine-tuning. %
Nor does a teacher's state help: pulled towards the model's own state at layer $t$ with the new fact in context, the bridge reads as the new entity in only 36--44\% of the window and composes 0.25, 0.32 and 0.25, no better than fine-tuning. %
The effect grows with $w$, does not depend on the target sentence, and no weight rescues the lenses (Appendix~\ref{app:design}). %

\subsection{Composition follows what the bridge reads as}\label{sec:distance}

The contract predicts that composition depends on what the bridge reads as while the second hop reads it, whichever method produced it.
For every trained model we take the bridge of each test question (not a training description) and ask, at every layer of the reading window fixed in Appendix~\ref{app:inject}, whether its hidden state is closer to the base model's reading of the new entity than of the old one.
The prediction holds question by question inside writers that never optimise the bridge: fine-tuning, CaKE and the logit-lens target answer 0.36--0.45 of the questions whose bridge reads as the new entity in most of the window, and \mbox{0.14--0.24} of the rest, on all three models. %
It holds across writers as well: four families, the weight sweep and the swapped targets fall on one curve (Figure~\ref{fig:axis}, right; Pearson $r=0.95$ over 18 writers on Qwen2.5, 0.84 and 0.89 over 14 on Llama-3 and Qwen3). %
Writers whose bridge reads as the old entity in about half of the window compose \mbox{0.18--0.36}, BRACE included. %
AlphaEdit never looks at the bridge but writes into the subject's early MLPs, where the model keeps its own facts; its bridge reads as the new entity in 68\% of the window, where \ours with $w=0.1$ also lands (67\%), and the two compose alike (0.53 and 0.51); \ours at $w=3$ reaches 84\% and 0.71. %
It does not depend on wording: on the 75\% of questions whose bridge phrase is never trained, \ours composes as often (Appendix~\ref{app:strict}).

\textbf{The model's own knowledge.} Its failures have the same cause: on 1{,}302 Wikidata facts Qwen2.5-7B answers, the logit lens ranks the object top-10 on 78\% of them where it is stated, and on at most 0.1\% at the bridge across layers \mbox{8--26}. %
When it knows both single hops, rewriting the first with \ours, adding no information, lifts chains with untrained second hops and wordings from 0.75 to 0.89; FT on the same text gives 0.57 (Appendix~\ref{app:own}). %
Hence facts written as read compose more often than the model's own.

\subsection{Design follows from where the question reads}\label{sec:design}

\textbf{Which token.} The aligned token must refer only to the bridge, so descriptions end in a noun phrase (Appendix~\ref{app:design}).
\textbf{Which layer.} Where the second hop finishes reading the bridge: aligned where reading starts (layer 10 on Qwen2.5), the bridge drifts back to 0.56 by layer 24 and \bacc is \mbox{0.57--0.59}; aligned where reading ends, it stays at 0.16 and \bacc is 0.62--0.71, because a trained write moves lower layers too. %
\textbf{Which neighbours.} Without $\mathcal{L}_{\text{keep}}$, same-relation retention falls below CaKE's; with it, retention exceeds CaKE's and rises with more anchors (Appendix~\ref{app:design}). %
\section{Related Work}\label{sec:related}

\paragraph{(I) Multi-hop knowledge editing.}
MQuAKE~\citep{mquake} and RippleEdits~\citep{rippleedits} showed that edits fail to propagate, and locate-then-edit methods~\citep{rome,memit,alphaedit} propagate poorly beyond the named subject~\citep{cake}; IFMET~\citep{ifmet} and ACE~\citep{ace} also edit deeper MLP layers or query--value neurons with multi-hop prompts: they change where the answer is written, not what the bridge becomes.
CaKE~\citep{cake} synthesizes circuit-aware data, \citet{kureason} self-distil on multi-hop questions, PropMEND~\citep{propmend} meta-learns propagation updates, and MCircKE~\citep{mcircke} edits reasoning circuits.
These methods supply more answers, or better answers; we change what is written.

\paragraph{(II) Latent multi-hop reasoning.}
Models perform the first hop latently and often fail at the second~\citep{yang2024latent}: bridges that resolve too late fail, and inference-time back-patching~\citep{hoppinglate} or middle-layer injections~\citep{memoryinjections,creme} restore composition.
\citet{balesni} named the two-hop curse and found it weaker when one hop is pre-trained, \citet{consistency} tie two-hop generalization to consistent entity representations across contexts, and BRACE~\citep{brace} observes mismatched bridge representations after knowledge injection and reuses shared descriptors across hops in training, a data-side route to the alignment that \ours writes directly (Table~\ref{tab:main}).
In models trained from scratch, aligning input and output bridge representations is necessary for two-hop generalization~\citep{identitybridge}.
These works observe that late or misaligned bridges break composition; we show that fine-tuning is what makes the bridge late, and say what it should be written as.

\paragraph{(III) Representation targets for training.}
\citet{balesni} added lens losses at the bridge~\citep{logitlens}; LRSD~\citep{lrsd} self-distils the subject's middle layers to its late layers; Patchscopes~\citep{patchscopes} read and patch representations across layers.
They pick targets from the model's output space or its own later layers; the contract's target is how the model reads.
\section{Conclusion}\label{sec:conclusion}

The two-hop curse is a write-time failure: fine-tuning writes a fact as an answer to be said; the second hop reads it as an entity.
Written the way the model reads it, a learned fact composes more often than the model's own knowledge.
Models that update their own weights will need to write what they learn the way they read it.

\raggedbottom %
\bibliography{refs}
\bibliographystyle{icml2025}

\newpage
\appendix
\onecolumn
\section{Implementation details}\label{app:impl}

\paragraph{Training.}
All writers follow CaKE's official protocol~\citep{cake}: one LoRA per edit (rank 8, $\alpha=32$, dropout 0.05, on the query, key, value, output, gate, up and down projections of every layer), 40 epochs, batch size 4, learning rate $10^{-4}$ with linear decay, bfloat16. %
FT trains on the edit sentence and six paraphrases; CaKE adds its released circuit-aware data; \ours adds $w\,(\mathcal{L}_{\text{read}}+\mathcal{L}_{\text{keep}})$ to FT with $w=3$.
LRSD uses its released configuration: the subject span at layer $L/2$ is pulled towards a stop-gradient copy of layer $3L/4$, with $\lambda=1$. %
The same weight is used for all models and benchmarks.
To train many edits at once, we stack the LoRA parameters of $K=8$ edits and route every sample only through its own edit's parameters; on a deterministic check this matches per-edit PEFT training to four decimals at the first step and within $3\cdot10^{-3}$ after 160 steps. %
One edit takes about 11 seconds on a single 85\,GB GPU with eight edits trained together. %

\paragraph{Read target and layer.}
The target $e_t(o)$ is the base model's hidden state at the last token of $o$ in ``This is about $o$'', at the output of layer $t$.
The descriptions are aligned at their last token: the subject's last token in ``the $N$ of $s$'' and the role noun in ``$s$'s $N$''.
$t$ is the layer at which the second hop finishes reading the bridge (Appendix~\ref{app:inject}): $t=24$ for Qwen2.5-7B-Instruct, $18$ for Llama-3-8B-Instruct and $26$ for Qwen3-8B.

\paragraph{Neighbours.}
For each edit, $\mathcal{L}_{\text{keep}}$ uses three other subjects of the edited relation, drawn from MQuAKE-CF and absent from the evaluation data; their descriptions are formed with the same role nouns.
On RippleEdits, neighbours are subjects of the same relation from RippleEdits edits whose entities do not occur in the evaluation data; on Balesni's benchmark, from the fictional people of the other sets.

\paragraph{Descriptions.}
Each model names the role of the answer for every relation from a five-shot prompt over relations that do not occur in the benchmarks:
\begin{quote}\small\ttfamily
Each fact template has a blank \{\} for the subject; the word that completes the template is the answer. Say what the answer is to the subject, as a short noun (1-3 words) that fits "the answer is the subject's \_\_".\\
Template: The official currency of \{\} is\\
The answer is the subject's: official currency; currency\\
\ldots{} (painted by $\to$ painter; artist, born in the year $\to$ birth year; year of birth, member of the band $\to$ band; music group, tallest mountain $\to$ tallest mountain; highest peak)\\
Template: \textrm{\textit{[template of the relation]}}\\
The answer is the subject's:
\end{quote}
The two nouns of the greedy answer give four descriptions per relation, ``the \textit{noun} of $s$'' and ``$s$'s \textit{noun}'' for each noun.

\paragraph{Evaluation.}
Generation is greedy.
MQuAKE answers are scored by alias substring match on two forms, the cloze (``Answer: '' followed by the multi-hop cloze, 10 tokens) and the question (``Question: \ldots\ Answer: The answer is'', 10 tokens); a case counts as correct if either form is, as in CaKE's protocol.
Locality counts, among same-relation neighbour questions and unrelated questions that the base model answers correctly, those still answered correctly after the edit.
RippleEdits uses the official test conditions (OR / AND over queries), 20 tokens and the ``Answer: '' prefix; only tests whose condition queries the base model answers correctly are kept, for all methods alike.
Balesni's benchmark uses its own chat format, system prompt and substring scoring, with 50 tokens.

\paragraph{Data.}
MQuAKE-CF-3k-v2: the first 100 cases form the development split.
Main-table two-hop cases: 300 for Qwen2.5, 100 for Llama-3 and Qwen3.
Two-hop-curse questions (an edited first hop followed by an unedited second hop): all 213 outside the development split, for every model.
RippleEdits: 273 popular-split edits with compositional tests; 572, 644 and 573 compositionality-I tests (Qwen2.5, Llama-3, Qwen3) after filtering.
Balesni's semi-synthetic benchmark: 18 sets of 20 facts and 1{,}380 two-hop questions.

\begin{table}[h]
\centering
\footnotesize
\setlength{\aboverulesep}{0pt}\setlength{\belowrulesep}{0pt}\renewcommand{\arraystretch}{1.08}
\setlength{\tabcolsep}{3pt}
\begin{tabular}{l *{6}{r}}
\toprule
\headrow & \multicolumn{2}{c}{\textbf{Qwen2.5}} & \multicolumn{2}{c}{\textbf{Llama-3}} & \multicolumn{2}{c}{\textbf{Qwen3}} \\
\cmidrule(lr){2-3}\cmidrule(lr){4-5}\cmidrule(lr){6-7}
\headrow \textbf{Method} & I & II & I & II & I & II \\
\midrule
FT & 0.30 & 0.76 & 0.26 & 0.80 & 0.28 & \underline{0.79} \\  %
CaKE & 0.31 & \textbf{0.96} & 0.29 & 0.80 & 0.29 & \textbf{1.00} \\  %
\oursrow \quad + \ours & \underline{0.57} & \textbf{0.96} & \underline{0.46} & \textbf{1.00} & \underline{0.53} & \textbf{1.00} \\  %
\oursrow \ours & \textbf{0.63} & \underline{0.92} & \textbf{0.49} & \underline{0.95} & \textbf{0.56} & 0.76 \\  %
\bottomrule
\end{tabular}
\caption{RippleEdits. CI: the new object is a bridge to a fact the model already knew (two-hop-curse questions). CII: the answer is the written object, reached through another subject. Questions: 572 / 644 / 573 (CI), 25 / 40 / 33 (CII). + \ours: CaKE trained with \ours added. \textbf{Bold}: best; \underline{underline}: second best.}
\label{tab:ripple}
\end{table}

\section{Reading window and layer choice}\label{app:inject}

On 200 (Qwen2.5, Qwen3) and 120 (Llama-3) two-hop chains of MQuAKE-CF that are not in MQuAKE-CF-3k-v2, not filtered for whether the model knows the second hop, we write the first-hop description into the second-hop prompt and, at its last token, replace the hidden state at layer $t$ with the base model's reading of the true bridge entity; each description is scored on one cloze.
Table~\ref{tab:inject} gives the plateau of the curves in Figure~\ref{fig:diag} (panel~2): where reading can start.
To find where it ends, we run the base model on the same kind of chains with its original facts (277, 246 and 242 chains it answers correctly) and, from layer $l$ onward, cut the attention of every later position to the bridge.
Cutting from early layers removes 40--48\% of the correct answers; cutting from the read end on keeps at least 90\% (Qwen2.5: 0.86 from layer 22, 0.99 from 24; Llama-3: 0.84 from 16, 0.95 from 18; Qwen3: 0.81 from 24, 0.93 from 26). %
The read end is the layer $t$ used for training; the rule was fixed before any test run.

\begin{table}[!htb]
\centering
\footnotesize
\setlength{\aboverulesep}{0pt}\setlength{\belowrulesep}{0pt}\renewcommand{\arraystretch}{1.1}
\setlength{\tabcolsep}{4.5pt}
\begin{tabular}{l c c c c c}
\toprule
\headrow \textbf{Model} & \textbf{None} & \textbf{Named} & \textbf{Plateau} & \textbf{Peak} & \textbf{End} $t$ \\
\midrule
Qwen2.5-7B & 0.06 & 0.34 & layers 2--10 & 0.20 & 24 \\ %
Llama-3-8B & 0.03 & 0.33 & layers 0--10 & 0.26 & 18 \\ %
Qwen3-8B & 0.04 & 0.22 & layers 6--12 & 0.20 & 26 \\ %
\bottomrule
\end{tabular}
\caption{Second-hop accuracy when the bridge is injected at one layer, on held-out chains, to locate the reading window; the rescue of \S2.3 is on the test questions. \emph{None}: no injection. \emph{Named}: the bridge entity written by name in the prompt instead of its description. Beyond the plateau, injection falls back towards no injection. \emph{End}: where reading ends, the shallowest layer from which cutting attention to the bridge keeps at least 90\% of the correct answers.}
\label{tab:inject}
\end{table}

\section{Design choices on held-out data}\label{app:design}

\paragraph{Which token.}
Descriptions must end on a token that refers only to the bridge.
Early hand-written descriptions ending in a verb or preposition (``the continent where $s$ is located'') pulled those words towards the new entity; on three-hop chains whose last hop is edited, the cloze then emitted ``located'' at the answer position.
Subject-final noun phrases and possessives remove the interference (0.69 $\to$ 0.92 on that subset). %

\paragraph{The column term.}
Table~\ref{tab:column} compares read-equivalence alone with the full objective, with hand-written descriptions.
The column term lifts both locality measures above CaKE's, and composition stays about twice CaKE's.

\begin{table}[!htb]
\centering
\footnotesize
\setlength{\aboverulesep}{0pt}\setlength{\belowrulesep}{0pt}\renewcommand{\arraystretch}{1.1}
\setlength{\tabcolsep}{3pt}
\begin{tabular}{l c c c c c c}
\toprule
\headrow & \multicolumn{3}{c}{\textbf{Qwen2.5 (300)}} & \multicolumn{3}{c}{\textbf{Llama-3 (100)}} \\
\headrow & \textbf{Comp.} & \textbf{Nbr} & \textbf{Unrel} & \textbf{Comp.} & \textbf{Nbr} & \textbf{Unrel} \\
\midrule
CaKE & 0.31 & 0.39 & 0.82 & 0.26 & 0.25 & 0.71 \\
$\mathcal{L}_{\text{read}}$ only & 0.63 & 0.32 & 0.78 & 0.52 & 0.08 & 0.53 \\
\oursrow $+\,\mathcal{L}_{\text{keep}}$ (\ours) & 0.68 & 0.47 & 0.86 & 0.48 & 0.30 & 0.73 \\
\bottomrule
\end{tabular}
\caption{Effect of the column term on MQuAKE two-hop cases. Comp.: questions that need composition (59 and 23). Nbr / Unrel: locality as in Table~\ref{tab:main}.} %
\label{tab:column}
\end{table}

\paragraph{Weight of the lens losses.}
The lens rows of Table~\ref{tab:main} use the weight of \ours ($w=3$). On 200 held-out chains (MQuAKE-CF two-hop chains outside the test set, Qwen2.5), the embed-lens loss gives 0.32--0.34 and the logit-lens loss 0.28--0.34 for every weight in $\{0.01, 0.1, 1, 3, 10\}$, so no weight rescues either target. %

\paragraph{Weight and target sentence of \ours.}
On the 213 two-hop-curse questions (Qwen2.5), \ours gives 0.30, 0.51, 0.63, 0.66, 0.71 and 0.69 at $w=0, 0.1, 0.3, 1, 3, 10$, and the median distance at layer $t$ falls with it (0.79, 0.49, 0.34, 0.23, 0.16, 0.15). Reading $o$ in ``This is about $o$'', ``The topic is $o$'' or ``We are talking about $o$'' gives 0.71, 0.68 and 0.71. %

\paragraph{Number of neighbours.}
The neighbour term anchors three other subjects of the edited relation per edit. More anchors keep more neighbours without lowering composition (Table~\ref{tab:nb}); the paper keeps the three fixed before the test runs.

\begin{table}[h]
\centering
\footnotesize
\setlength{\tabcolsep}{5pt}
\begin{tabular}{l rrrr}
\toprule
\headrow \textbf{Neighbours per edit} & Any & Composition & Nbr & Unrel \\
\midrule
3 (used) & 0.77 & 0.56 & 0.64 & 0.88 \\  %
10 & 0.75 & 0.67 & 0.76 & 0.91 \\  %
20 & 0.72 & 0.63 & 0.77 & 0.90 \\  %
30 & 0.71 & 0.59 & 0.80 & 0.91 \\  %
\bottomrule
\end{tabular}
\caption{Number of anchored neighbours on the development split (Qwen2.5, 100 cases; composition: the 27 two-hop-curse cases).}
\label{tab:nb}
\end{table}

\paragraph{Generated vs.\ hand-written descriptions.}
On the development split, descriptions generated by the edited model match hand-written ones (Qwen2.5 0.80 vs.\ 0.76, Llama-3 0.76 vs.\ 0.78); the paper uses generated descriptions throughout. %

\section{AlphaEdit}\label{app:alphaedit}

We run the official AlphaEdit implementation~\citep{alphaedit} with our evaluation. The null-space projection is estimated from 20{,}000 Wikipedia articles; Llama-3 uses the released hyper-parameters, and Qwen2.5 and Qwen3 use those of EasyEdit for Qwen2.5 (Qwen3 with the value loss at its last layer). Each case is edited independently and the weights are restored afterwards, as for every other writer. %

\begin{table}[!htb]
\centering
\footnotesize
\setlength{\aboverulesep}{0pt}\setlength{\belowrulesep}{0pt}\renewcommand{\arraystretch}{1.1}
\setlength{\tabcolsep}{1.3pt}
\begin{tabular}{l r r r >{\hspace{3pt}}r r >{\hspace{3pt}}r r r r >{\hspace{3pt}}r r >{\hspace{3pt}}r r r r >{\hspace{3pt}}r r >{\hspace{3pt}}r}
\toprule
\headrow & \multicolumn{6}{c}{\textbf{Qwen2.5-7B}} & \multicolumn{6}{c}{\textbf{Llama-3-8B}} & \multicolumn{6}{c}{\textbf{Qwen3-8B}} \\
\cmidrule(lr){2-7}\cmidrule(lr){8-13}\cmidrule(lr){14-19}
\headrow \textbf{Method} & \multicolumn{1}{c}{M-Acc} & \multicolumn{1}{c}{\bacc} & \multicolumn{1}{c}{CR} & \multicolumn{1}{c}{Nbr} & \multicolumn{1}{c}{Unrel} & \multicolumn{1}{c}{Avg} & \multicolumn{1}{c}{M-Acc} & \multicolumn{1}{c}{\bacc} & \multicolumn{1}{c}{CR} & \multicolumn{1}{c}{Nbr} & \multicolumn{1}{c}{Unrel} & \multicolumn{1}{c}{Avg} & \multicolumn{1}{c}{M-Acc} & \multicolumn{1}{c}{\bacc} & \multicolumn{1}{c}{CR} & \multicolumn{1}{c}{Nbr} & \multicolumn{1}{c}{Unrel} & \multicolumn{1}{c}{Avg} \\
\midrule
AlphaEdit & 0.19 & 0.53 & 0.61 & \textbf{0.98} & \textbf{0.99} & 0.67 & 0.10 & 0.46 & 0.60 & \textbf{0.99} & \textbf{0.99} & 0.64 & 0.14 & 0.48 & 0.66 & \textbf{0.93} & \textbf{0.92} & 0.62 \\  %
\oursrow \ours & \textbf{0.83} & \textbf{0.71} & \textbf{0.81} & 0.66 & 0.92 & \textbf{0.78} & \textbf{0.83} & \textbf{0.62} & \textbf{0.68} & 0.46 & 0.79 & \textbf{0.68} & \textbf{0.79} & \textbf{0.63} & \textbf{0.78} & 0.62 & \textbf{0.92} & \textbf{0.74} \\  %
\bottomrule
\end{tabular}
\caption{AlphaEdit and \ours on MQuAKE, with the columns of Table~\ref{tab:main}. \textbf{Bold}: the better of the two.}
\label{tab:ae}
\end{table}

AlphaEdit's edit is keyed to the subject named in the prompt. Splitting the two-hop cases by which hop is edited (Table~\ref{tab:ae_split}): when only the first hop is edited, the edited subject is named in the question and AlphaEdit answers 0.30--0.58; when the second hop is edited, its subject is the bridge, which the question never names, and AlphaEdit answers 0.02--0.12, where \ours answers 0.73--0.94.

\begin{table}[h]
\centering
\footnotesize
\setlength{\tabcolsep}{4pt}
\begin{tabular}{l rr rr rr}
\toprule
\headrow & \multicolumn{2}{c}{\textbf{Qwen2.5-7B}} & \multicolumn{2}{c}{\textbf{Llama-3-8B}} & \multicolumn{2}{c}{\textbf{Qwen3-8B}} \\
\headrow \textbf{Edited hop} & AlphaEdit & \ours & AlphaEdit & \ours & AlphaEdit & \ours \\
\midrule
First only & 0.58 & 0.80 & 0.30 & 0.61 & 0.39 & 0.70 \\  %
Second only & 0.09 & 0.82 & 0.06 & 0.94 & 0.12 & 0.73 \\  %
Both & 0.10 & 0.85 & 0.02 & 0.86 & 0.02 & 0.89 \\  %
\bottomrule
\end{tabular}
\caption{Multi-hop accuracy on the main-table two-hop cases, split by which hop is edited.}
\label{tab:ae_split}
\end{table}

\section{BRACE}\label{app:brace}

BRACE~\citep{brace} injects two counterfactual hops and writes the bridge in both with a descriptor from one pool of eight type-consistent noun phrases, rotating determiners (``the'', ``that'', ``the famous'', none); the tail is never decorated.
In MQuAKE an edit can fall on either hop and the other hop is the model's own knowledge, so we decorate whichever entity of the edited fact is a bridge (the new object unless the last hop is edited, the subject unless the first hop is edited) and add about six known facts about the bridge, decorated from the same pool, on the side of the hop that is not trained.
Pools and facts are generated by DeepSeek-V4.1-Flash; training follows the protocol of all writers. %
BRACE's tails are counterfactual, so real-world descriptors cannot reveal them; ours are real, and a descriptor often names the answer (``the East Asian island nation Taiwan'' for a question whose answer is Asia).
Following BRACE's audit that the pool contain no tail, we drop every descriptor or fact that contains the answer, an alias or a word stem of it; the answer then occurs as a word in the training text on the same 31 of the 213 questions as in FT's, all from the released edits and paraphrases (``which Asian country'', ``University of Paris''). Without these 31 questions, B-Acc is 0.66 / 0.57 / 0.57 for \ours, 0.21 / 0.20 / 0.22 for FT and 0.53 / 0.50 / 0.50 for the model's own knowledge. %

A model trained this way names a descriptor before every entity it answers, undecorated tails included (``that the author of Coraline belongs to that sheep-farming archipelago New Zealand''), and the 10-token budget can end such an answer before the name. %
BRACE generates 64 tokens; longer outputs run on into further entities, so Table~\ref{tab:main} scores the first sentence of each. Table~\ref{tab:brace} gives both scores.

\begin{table}[h]
\centering
\footnotesize
\setlength{\aboverulesep}{0pt}\setlength{\belowrulesep}{0pt}\renewcommand{\arraystretch}{1.08}
\setlength{\tabcolsep}{3pt}
\begin{tabular}{l *{9}{r}}
\toprule
\headrow & \multicolumn{3}{c}{\textbf{Qwen2.5-7B}} & \multicolumn{3}{c}{\textbf{Llama-3-8B}} & \multicolumn{3}{c}{\textbf{Qwen3-8B}} \\
\cmidrule(lr){2-4}\cmidrule(lr){5-7}\cmidrule(lr){8-10}
\headrow \textbf{Scoring} & M-Acc & \bacc & CR & M-Acc & \bacc & CR & M-Acc & \bacc & CR \\
\midrule
BRACE, first 10 tokens & 0.45 & 0.25 & 0.31 & 0.49 & 0.24 & 0.26 & 0.55 & 0.26 & 0.33 \\  %
BRACE, first sentence of 64 & 0.59 & 0.36 & 0.47 & 0.70 & 0.37 & 0.43 & 0.67 & 0.36 & 0.46 \\  %
\oursrow \ours, first 10 tokens & 0.83 & 0.71 & 0.81 & 0.83 & 0.62 & 0.68 & 0.79 & 0.63 & 0.78 \\  %
\bottomrule
\end{tabular}
\caption{BRACE under the 10-token budget of all other rows and under its own 64-token generation, scored on the first sentence.}
\label{tab:brace}
\end{table}

\section{Intervals and paired tests}\label{app:ci}

Table~\ref{tab:ci} gives each main-table entry with its 95\% interval, a percentile bootstrap over edit cases (10{,}000 resamples; locality resampled by case). Paired sign tests over cases compare \ours with every baseline of Table~\ref{tab:main}. On all three models \ours is higher on \bacc and CR against every baseline ($p<10^{-3}$; on \bacc, 60--109 wins against 0--10 losses) and on Nbr ($p<0.05$), and on \bacc it exceeds the base model's own knowledge on the same chains ($p=10^{-4}$, $0.013$, $0.006$; 57--22, 60--35, 48--24). On Qwen2.5, with 300 cases, its M-Acc and Unrel are higher than every baseline ($p<0.01$). %

\begin{table}[h]
\centering
\footnotesize
\setlength{\aboverulesep}{0pt}\setlength{\belowrulesep}{0pt}\renewcommand{\arraystretch}{1.08}
\setlength{\tabcolsep}{4pt}
\begin{tabular}{l *{5}{r}}
\toprule
\headrow \textbf{Method} & \multicolumn{1}{c}{M-Acc} & \multicolumn{1}{c}{\bacc} & \multicolumn{1}{c}{CR} & \multicolumn{1}{c}{Nbr} & \multicolumn{1}{c}{Unrel} \\
\midrule
\multicolumn{6}{l}{\textbf{Qwen2.5-7B}} \\
\textcolor{refc}{\textit{Own knowledge}} & -- & \textcolor{refc}{0.54 {\scriptsize[0.48, 0.61]}} & -- & -- & -- \\
FT & 0.71 {\scriptsize[0.65, 0.76]} & 0.30 {\scriptsize[0.24, 0.36]} & 0.37 {\scriptsize[0.28, 0.46]} & 0.42 {\scriptsize[0.38, 0.46]} & 0.83 {\scriptsize[0.80, 0.86]} \\  %
CaKE & 0.75 {\scriptsize[0.70, 0.80]} & 0.27 {\scriptsize[0.22, 0.33]} & 0.32 {\scriptsize[0.23, 0.41]} & 0.39 {\scriptsize[0.35, 0.44]} & 0.82 {\scriptsize[0.78, 0.85]} \\  %
\oursrow \quad + \ours & 0.87 {\scriptsize[0.83, 0.91]} & 0.64 {\scriptsize[0.57, 0.70]} & 0.75 {\scriptsize[0.67, 0.83]} & 0.63 {\scriptsize[0.59, 0.67]} & 0.90 {\scriptsize[0.88, 0.92]} \\  %
Self-distillation & 0.70 {\scriptsize[0.65, 0.76]} & 0.34 {\scriptsize[0.28, 0.41]} & 0.42 {\scriptsize[0.34, 0.51]} & 0.43 {\scriptsize[0.38, 0.47]} & 0.79 {\scriptsize[0.76, 0.82]} \\  %
BRACE$^\ddagger$ & 0.59 {\scriptsize[0.53, 0.65]} & 0.36 {\scriptsize[0.29, 0.42]} & 0.47 {\scriptsize[0.38, 0.56]} & 0.48 {\scriptsize[0.43, 0.52]} & 0.81 {\scriptsize[0.78, 0.84]} \\  %
Embed-lens & 0.64 {\scriptsize[0.59, 0.70]} & 0.26 {\scriptsize[0.21, 0.32]} & 0.34 {\scriptsize[0.25, 0.42]} & 0.31 {\scriptsize[0.27, 0.35]} & 0.75 {\scriptsize[0.72, 0.79]} \\  %
Logit-lens & 0.72 {\scriptsize[0.67, 0.77]} & 0.29 {\scriptsize[0.23, 0.35]} & 0.37 {\scriptsize[0.28, 0.46]} & 0.41 {\scriptsize[0.37, 0.45]} & 0.82 {\scriptsize[0.79, 0.85]} \\  %
LRSD & 0.74 {\scriptsize[0.69, 0.79]} & 0.23 {\scriptsize[0.17, 0.28]} & 0.27 {\scriptsize[0.19, 0.34]} & 0.16 {\scriptsize[0.13, 0.20]} & 0.55 {\scriptsize[0.51, 0.59]} \\  %
\oursrow \ours & 0.83 {\scriptsize[0.79, 0.87]} & 0.71 {\scriptsize[0.65, 0.77]} & 0.81 {\scriptsize[0.73, 0.88]} & 0.66 {\scriptsize[0.62, 0.70]} & 0.92 {\scriptsize[0.90, 0.94]} \\  %
\midrule
\multicolumn{6}{l}{\textbf{Llama-3-8B}} \\
\textcolor{refc}{\textit{Own knowledge}} & -- & \textcolor{refc}{0.51 {\scriptsize[0.44, 0.57]}} & -- & -- & -- \\
FT & 0.68 {\scriptsize[0.59, 0.77]} & 0.30 {\scriptsize[0.24, 0.36]} & 0.35 {\scriptsize[0.27, 0.44]} & 0.26 {\scriptsize[0.19, 0.32]} & 0.65 {\scriptsize[0.58, 0.71]} \\  %
CaKE & 0.75 {\scriptsize[0.66, 0.83]} & 0.34 {\scriptsize[0.28, 0.41]} & 0.38 {\scriptsize[0.29, 0.47]} & 0.25 {\scriptsize[0.19, 0.31]} & 0.71 {\scriptsize[0.66, 0.76]} \\  %
\oursrow \quad + \ours & 0.79 {\scriptsize[0.71, 0.87]} & 0.63 {\scriptsize[0.57, 0.69]} & 0.71 {\scriptsize[0.63, 0.80]} & 0.45 {\scriptsize[0.37, 0.52]} & 0.81 {\scriptsize[0.77, 0.86]} \\  %
Self-distillation & 0.73 {\scriptsize[0.64, 0.81]} & 0.38 {\scriptsize[0.31, 0.44]} & 0.44 {\scriptsize[0.35, 0.54]} & 0.33 {\scriptsize[0.26, 0.41]} & 0.73 {\scriptsize[0.67, 0.79]} \\  %
BRACE$^\ddagger$ & 0.70 {\scriptsize[0.61, 0.79]} & 0.37 {\scriptsize[0.31, 0.44]} & 0.43 {\scriptsize[0.33, 0.52]} & 0.37 {\scriptsize[0.30, 0.45]} & 0.70 {\scriptsize[0.64, 0.76]} \\  %
Embed-lens & 0.60 {\scriptsize[0.50, 0.69]} & 0.15 {\scriptsize[0.11, 0.21]} & 0.17 {\scriptsize[0.10, 0.24]} & 0.09 {\scriptsize[0.04, 0.14]} & 0.54 {\scriptsize[0.47, 0.60]} \\  %
Logit-lens & 0.68 {\scriptsize[0.59, 0.77]} & 0.26 {\scriptsize[0.20, 0.32]} & 0.31 {\scriptsize[0.22, 0.39]} & 0.16 {\scriptsize[0.10, 0.22]} & 0.57 {\scriptsize[0.50, 0.64]} \\  %
LRSD & 0.67 {\scriptsize[0.58, 0.76]} & 0.11 {\scriptsize[0.07, 0.15]} & 0.12 {\scriptsize[0.06, 0.19]} & 0.03 {\scriptsize[0.00, 0.06]} & 0.07 {\scriptsize[0.04, 0.10]} \\  %
\oursrow \ours & 0.83 {\scriptsize[0.75, 0.90]} & 0.62 {\scriptsize[0.56, 0.69]} & 0.68 {\scriptsize[0.58, 0.76]} & 0.46 {\scriptsize[0.38, 0.54]} & 0.79 {\scriptsize[0.74, 0.84]} \\  %
\midrule
\multicolumn{6}{l}{\textbf{Qwen3-8B}} \\
\textcolor{refc}{\textit{Own knowledge}} & -- & \textcolor{refc}{0.52 {\scriptsize[0.45, 0.58]}} & -- & -- & -- \\
FT & 0.67 {\scriptsize[0.58, 0.76]} & 0.30 {\scriptsize[0.23, 0.36]} & 0.37 {\scriptsize[0.28, 0.46]} & 0.33 {\scriptsize[0.25, 0.41]} & 0.75 {\scriptsize[0.68, 0.82]} \\  %
CaKE & 0.73 {\scriptsize[0.64, 0.82]} & 0.31 {\scriptsize[0.24, 0.37]} & 0.37 {\scriptsize[0.28, 0.46]} & 0.31 {\scriptsize[0.23, 0.38]} & 0.75 {\scriptsize[0.69, 0.81]} \\  %
\oursrow \quad + \ours & 0.80 {\scriptsize[0.72, 0.87]} & 0.57 {\scriptsize[0.50, 0.63]} & 0.69 {\scriptsize[0.60, 0.77]} & 0.56 {\scriptsize[0.48, 0.63]} & 0.87 {\scriptsize[0.83, 0.91]} \\  %
Self-distillation & 0.75 {\scriptsize[0.66, 0.83]} & 0.38 {\scriptsize[0.31, 0.44]} & 0.45 {\scriptsize[0.36, 0.55]} & 0.37 {\scriptsize[0.29, 0.45]} & 0.73 {\scriptsize[0.66, 0.80]} \\  %
BRACE$^\ddagger$ & 0.67 {\scriptsize[0.58, 0.76]} & 0.36 {\scriptsize[0.29, 0.42]} & 0.46 {\scriptsize[0.37, 0.55]} & 0.39 {\scriptsize[0.32, 0.47]} & 0.79 {\scriptsize[0.73, 0.85]} \\  %
Embed-lens & 0.59 {\scriptsize[0.49, 0.69]} & 0.20 {\scriptsize[0.15, 0.26]} & 0.28 {\scriptsize[0.20, 0.36]} & 0.22 {\scriptsize[0.16, 0.29]} & 0.64 {\scriptsize[0.57, 0.70]} \\  %
Logit-lens & 0.64 {\scriptsize[0.55, 0.73]} & 0.30 {\scriptsize[0.23, 0.36]} & 0.38 {\scriptsize[0.29, 0.47]} & 0.38 {\scriptsize[0.30, 0.46]} & 0.80 {\scriptsize[0.74, 0.85]} \\  %
LRSD & 0.66 {\scriptsize[0.57, 0.75]} & 0.25 {\scriptsize[0.20, 0.31]} & 0.34 {\scriptsize[0.25, 0.43]} & 0.14 {\scriptsize[0.08, 0.20]} & 0.47 {\scriptsize[0.40, 0.55]} \\  %
\oursrow \ours & 0.79 {\scriptsize[0.71, 0.87]} & 0.63 {\scriptsize[0.56, 0.69]} & 0.78 {\scriptsize[0.70, 0.85]} & 0.62 {\scriptsize[0.55, 0.70]} & 0.92 {\scriptsize[0.88, 0.95]} \\  %
\bottomrule
\end{tabular}
\caption{Main-table results with 95\% intervals (bootstrap over edit cases).}
\label{tab:ci}
\end{table}

\section{Strict matching and wording overlap}\label{app:strict}

\paragraph{Strict matching.} Our scoring follows CaKE: an alias of the answer anywhere in the greedy output, on the cloze or the question form. The strict score also requires the answer in the first clause of the output and rejects any output that names the bridge entity or the old answer. All writers drop, and the gap to the baselines widens (Table~\ref{tab:strict}).

\paragraph{Wording overlap.} A training description of \ours occurs verbatim in the test question for about a quarter of the 213 questions (53 / 59 / 65). On the other three quarters, which include clause forms the descriptions never take (``the official language of the country where Yulia Lipnitskaya holds citizenship'', Table~\ref{tab:examples}), \ours composes as often as on the overlapping ones (Table~\ref{tab:strict}), and on Qwen2.5 the share of questions whose bridge reads closer to the new entity than to the old one at layer $t$ keeps the same order across writers: 0.27 before the edit, 0.42 FT, 0.44 CaKE, 0.54 logit lens, 0.72 AlphaEdit, 1.00 \ours (all 213: 0.23, 0.37, 0.38, 0.49, 0.71, 1.00). %

\begin{table}[h]
\centering
\footnotesize
\setlength{\tabcolsep}{3.5pt}
\begin{tabular}{l rrrr rrrr rrrr}
\toprule
\headrow & \multicolumn{4}{c}{\textbf{Qwen2.5-7B}} & \multicolumn{4}{c}{\textbf{Llama-3-8B}} & \multicolumn{4}{c}{\textbf{Qwen3-8B}} \\
\headrow \textbf{Writer} & \bacc & Strict & Overlap & None & \bacc & Strict & Overlap & None & \bacc & Strict & Overlap & None \\
\midrule
\textcolor{refc}{\textit{Own knowledge}} & \textcolor{refc}{0.54} & \textcolor{refc}{0.35} & -- & -- & \textcolor{refc}{0.51} & \textcolor{refc}{0.30} & -- & -- & \textcolor{refc}{0.52} & \textcolor{refc}{0.37} & -- & -- \\  %
\midrule
FT & 0.30 & 0.15 & 0.26 & 0.31 & 0.30 & 0.11 & 0.27 & 0.31 & 0.30 & 0.21 & 0.22 & 0.33 \\  %
CaKE & 0.27 & 0.14 & 0.21 & 0.29 & 0.34 & 0.18 & 0.32 & 0.35 & 0.31 & 0.20 & 0.25 & 0.33 \\
Self-distillation & 0.34 & 0.20 & 0.26 & 0.37 & 0.38 & 0.17 & 0.39 & 0.37 & 0.38 & 0.27 & 0.38 & 0.37 \\
Embed-lens & 0.26 & 0.14 & 0.26 & 0.26 & 0.15 & 0.05 & 0.10 & 0.18 & 0.20 & 0.12 & 0.17 & 0.22 \\
Logit-lens & 0.29 & 0.17 & 0.23 & 0.31 & 0.26 & 0.13 & 0.17 & 0.29 & 0.30 & 0.23 & 0.23 & 0.32 \\
LRSD & 0.23 & 0.15 & 0.11 & 0.26 & 0.11 & 0.03 & 0.10 & 0.12 & 0.25 & 0.17 & 0.14 & 0.30 \\
\oursrow \ours & \textbf{0.71} & \textbf{0.50} & \textbf{0.75} & \textbf{0.69} & 0.62 & 0.45 & \textbf{0.63} & 0.62 & \textbf{0.63} & \textbf{0.53} & \textbf{0.62} & \textbf{0.64} \\
\oursrow CaKE + \ours & 0.64 & 0.48 & 0.62 & 0.64 & \textbf{0.63} & \textbf{0.46} & \textbf{0.63} & \textbf{0.64} & 0.57 & 0.46 & \textbf{0.62} & 0.55 \\
\bottomrule
\end{tabular}
\caption{Two-hop-curse questions under the default and the strict score, and split by whether a training description occurs verbatim in the question (Overlap) or not (None). \textbf{Bold}: best.}
\label{tab:strict}
\end{table}

\section{Balesni's semi-synthetic benchmark}\label{app:balesni}

Each of the 18 sets writes 20 facts ``fictional person $\to$ real entity'' (favourite, workplace or ownership relations) jointly into one LoRA, using the benchmark's own chat-format training conversations.
Descriptions follow the training relation (``$s$'s favorite T'', ``the T where $s$ works'', ``the T that $s$ owns''), without looking at test questions.
The two auxiliary losses of \citet{balesni}, embed-lens ($1-\cos$) and logit-lens (cross-entropy), are applied at the same position and layer as \ours, with their original coefficients (0.1 and 0.01) and with ours ($w=3$).
Table~\ref{tab:balfull} gives all runs; the strict score additionally requires that the answer does not contain the bridge entity's name.

\begin{table}[!htb]
\centering
\footnotesize
\setlength{\aboverulesep}{0pt}\setlength{\belowrulesep}{0pt}\renewcommand{\arraystretch}{1.1}
\setlength{\tabcolsep}{3.5pt}
\begin{tabular}{l c c c c}
\toprule
\headrow & \multicolumn{2}{c}{\textbf{Llama-3-8B}} & \multicolumn{2}{c}{\textbf{Qwen2.5-7B}} \\
\headrow \textbf{Writer} & \textbf{Two-hop} & \textbf{Strict} & \textbf{Two-hop} & \textbf{Strict} \\
\midrule
\textcolor{refc}{\textit{Ceiling (second hop)}} & \textcolor{refc}{0.67} & -- & \textcolor{refc}{0.66} & -- \\
Before the edit & 0.08 & 0.08 & 0.09 & 0.09 \\
FT & 0.17 & 0.15 & 0.11 & 0.10 \\
Embed-lens, $w=0.1$ & 0.17 & 0.14 & 0.09 & 0.09 \\
Embed-lens, $w=3$ & 0.16 & 0.15 & 0.11 & 0.11 \\
Logit-lens, $w=0.01$ & 0.22 & 0.19 & 0.14 & 0.13 \\
Logit-lens, $w=3$ & 0.20 & 0.18 & 0.20 & 0.18 \\
\oursrow \ours & \textbf{0.49} & \textbf{0.47} & \textbf{0.41} & \textbf{0.40} \\
\bottomrule
\end{tabular}
\caption{Balesni's semi-synthetic benchmark, 1{,}380 two-hop questions without chain-of-thought. The first hop is recalled perfectly after every write. \ours wins 330--473 questions and loses 16--43 against every other writer.} %
\label{tab:balfull}
\end{table}

\section{Examples}\label{app:examples}

Table~\ref{tab:examples} shows two-hop-curse questions from the test set (Qwen2.5) on which fine-tuning and CaKE fail and \ours answers.
The failures take two forms that the read--write contract predicts: the model says the written object itself where the bridge should be read (``Latvia'', ``England''), or it reads the old bridge (``Rome'' for gelato's old country, ``Italy'' for the creator of Hannibal Lecter).

\begin{table}[!htb]
\centering
\footnotesize
\setlength{\aboverulesep}{0pt}\setlength{\belowrulesep}{0pt}\renewcommand{\arraystretch}{1.15}
\setlength{\tabcolsep}{4pt}
\begin{tabular}{>{\raggedright\arraybackslash}p{4.3cm} >{\raggedright\arraybackslash}p{4.6cm} >{\raggedright\arraybackslash}p{1.9cm} >{\raggedright\arraybackslash}p{1.9cm} >{\raggedright\arraybackslash}p{2.3cm}}
\toprule
\headrow \textbf{Edit} & \textbf{Two-hop question} & \textbf{FT} & \textbf{CaKE} & \textbf{\ours} \\
\midrule
Yulia Lipnitskaya is a citizen of Latvia & The official language of the country where Yulia Lipnitskaya holds citizenship is & Latvia. & Russian. & \textcolor{readc}{Latvian.} \\
The Fireballs was created in the country of England & The head of state of the country where The Fireballs originated is & England. & England. & \textcolor{readc}{Elizabeth II.} \\
Gelato was created in the country of Portugal & The capital of the country where gelato originated is & Rome. & Rome. & \textcolor{readc}{Lisbon.} \\
Hannibal Lecter was created by Spede Pasanen & The country of origin of the creator of Hannibal Lecter is & Italy. & Italy. & \textcolor{readc}{Finland.} \\
Iwakura Tomomi is a citizen of New Zealand & The head of government of the country where Iwakura Tomomi holds citizenship is & also a citizen of New Zealand. & Andrew Scott Brown. & \textcolor{readc}{Jacinda Ardern.} \\
The author of The Comedy of Errors is Brian O'Nolan & The institution where the author of ``The Comedy of Errors'' received education is & Oxford University. & University of Padua. & \textcolor{readc}{University College Dublin.} \\
\bottomrule
\end{tabular}
\caption{Greedy cloze outputs on two-hop-curse questions (MQuAKE, Qwen2.5-7B).} %
\label{tab:examples}
\end{table}

\section{Controlled RippleEdit}\label{app:cre}

On Controlled RippleEdit~\citep{propmend}, a fictional subject is written with relations to three real entities, and every question asks for a known property of one of them, reached through the fictional subject.
Writing as read lifts accuracy from 0.08 to 0.34 on \texttt{test\_id} and from 0.04 to 0.21 on \texttt{test\_ood} (Qwen2.5-7B), on 101 wins vs.\ 12 losses and 16~vs.~1. %

\section{Injecting a new relation}\label{app:synthetic}

On a synthetic relation (``holiday destination'') written into Qwen3-4B, injection at middle layers (8--16) lifts the second hop from 0.10 to 0.80 and injection from layer 24 on reaches at most 0.17; %
read-equivalent writing lifts the language second hop from 0.47 to 0.82, the injection ceiling; with the same position and layer, the embedding and logit-lens targets reach 0.60 and 0.57. %

\section{Rewriting the model's own knowledge}\label{app:own}

On the 213 two-hop-curse chains built from the model's own knowledge (the \textit{Own knowledge} row of Table~\ref{tab:main}), the base model fails about half of the time. We restore each first-hop fact to its true object and rewrite it with \ours, adding no information: the paraphrases, descriptions, layers and weight of an edit, one case at a time (Qwen2.5-7B); FT on the same paraphrases is the control.
Besides the original chain, each case is tested on two-hop questions that keep its bridge and replace the second hop by a relation never seen in training (365 questions on 139 cases, built from MQuAKE facts about the bridge), with the bridge described either by a description of \ours or by one never trained (``the \emph{relation} associated with $s$''). %
Trained models answer tersely while the base model often reasons aloud, so every row here prefixes the same four demonstrations of direct answers and is scored by an LLM judge (DeepSeek-V4.1-Flash): correct when the first answer given is right and the bridge is not named before it. %

\begin{table}[h]
\centering
\footnotesize
\setlength{\tabcolsep}{4pt}
\begin{tabular}{l rrr r rr}
\toprule
\headrow & \multicolumn{3}{c}{\textbf{Unseen second hop}} & & \multicolumn{2}{c}{\textbf{Locality}} \\
\headrow \textbf{Writer} & Untrained desc. & Trained desc. & Base wrong & Original chain & Nbr & Unrel \\
\midrule
\textcolor{refc}{\textit{Own knowledge}} & \textcolor{refc}{0.75} & \textcolor{refc}{0.67} & \textcolor{refc}{0.00} & \textcolor{refc}{0.54} & \textcolor{refc}{0.89} & \textcolor{refc}{0.75} \\  %
FT & 0.57 & 0.51 & 0.18 & 0.46 & 0.77 & 0.69 \\
\oursrow \ours & \textbf{0.89} & \textbf{0.88} & \textbf{0.74} & \textbf{0.67} & 0.86 & 0.73 \\
\bottomrule
\end{tabular}
\caption{Rewriting a known fact with \ours (Qwen2.5-7B). Unseen second hop: questions whose two single hops the base model answers (203 per description); Base wrong: those the base model fails (untrained description, 50).}
\label{tab:own}
\end{table}

Rewriting one fact lifts the chains through it, for relations and wordings it was never trained on (Table~\ref{tab:own}): with the untrained description, 0.75 $\to$ 0.89 (37 wins vs.\ 10 losses), fixing three quarters of the questions the base model fails, while FT on the same text drops to 0.57 (\ours wins 68 vs.\ 3). When the head entity is swapped for one with another bridge, the base model still gives the original answer on 7.9\% of the questions; without them, 0.74 $\to$ 0.88. %
Written jointly into one LoRA, the 213 facts give 0.75 $\to$ 0.81 (FT 0.62). %

The base model's own knowledge fails for the reason of Section~\ref{sec:diagnosis}: it is stored as an answer to say. On 1{,}302 Wikidata facts ``work $\to$ author'' that Qwen2.5-7B answers, the logit lens ranks the author among the top 10 tokens on 78\% of them at the end of ``$s$ was written by'' (layer 26), and on at most 0.1\% at the bridge of ``The person who wrote $s$ died in'' at every layer from 8 to 26. Replacing the bridge state at layer 8 by the model's reading of the author raises two-hop accuracy from 0.27 to 0.39 (758 cloze questions, greedy 12 tokens). %
The repair is per fact: a linear map from bridge states to reading states fitted on half of the authors does not help on the other half (0.23--0.27), and rewriting 120 facts of a relation leaves its other facts unchanged (0.54 before, 0.56 after). %

\end{document}